\documentclass[sigconf,nonacm,screen]{acmart}

\usepackage{graphicx}
\graphicspath{{figures/}}
\usepackage{booktabs}
\usepackage{multirow}
\usepackage[dvipsnames]{xcolor}
\AtBeginDocument{
    \hypersetup{
        colorlinks=true,       
        linkcolor=Red,        
        filecolor=magenta,     
        urlcolor=magenta,
        citecolor=cyan, 
        pdftitle={From Retrieved Context to Runtime Control: Adaptive Compression for Edge-based RAG},
        pdfpagemode=FullScreen,
    }
}

\usepackage{xcolor}

\newif\ifcollabnotes
\collabnotesfalse

\setcopyright{none}
\renewcommand\footnotetextcopyrightpermission[1]{}

\title{From Retrieved Context to Runtime Control: Adaptive Compression for Edge-based RAG}
\titlenote{Accepted to appear in the Proceedings of the ACM AI Leadership Summit 2026.}
\author{Zlatan Feric}
\authornote{Both authors contributed equally to this research.}
\affiliation{%
  \institution{Northeastern University}
  \city{Boston}
  \state{Massachusetts}
  \country{USA}}
\email{feric.z@northeastern.edu}

\author{Amir Taherin}
\authornotemark[2]
\affiliation{%
  \institution{Northeastern University}
  \city{Boston}
  \state{Massachusetts}
  \country{USA}}
\email{taherin.a@northeastern.edu}

\author{Yanzhi Wang}
\affiliation{%
  \institution{Northeastern University}
  \city{Boston}
  \state{Massachusetts}
  \country{USA}}
\email{yanz.wang@northeastern.edu}

\author{David Kaeli}
\affiliation{%
  \institution{Northeastern University}
  \city{Boston}
  \state{Massachusetts}
  \country{USA}}
\email{d.kaeli@northeastern.edu}

\begin{document}

\begin{abstract}
Retrieval-augmented generation (RAG) improves language-model responses by grounding generation in external passages, which comes with overhead: retrieved context lengthens the prompt, increasing prefill work, KV-cache footprint, memory traffic, latency, and energy. Context compression offers a natural remedy by pruning retrieved text before generation. However, state-of-the-art context-compression methods are typically used with a fixed compression budget, or with the rate selected offline and then applied at inference time. This static view ignores both workload variation and the live state of the edge device. On an edge SoC, compression is not free: the compressor itself runs on the same SoC and consumes latency and energy that can offset any generation savings. 

This paper proposes a vision for telemetry-informed adaptive compression in edge RAG, grounded in experimental evidence. We characterize the compression tradeoff on the NVIDIA Jetson AGX Thor using Llama and Qwen generators, Natural Questions and HotpotQA datasets, and LLMLingua-2 compression. Our measurements show that generation dominates the RAG budget for larger models, reaching roughly 90\% of per-query latency and 91\% of GPU energy for 7B--8B generators. Exploring the impact of the compression rate reveals an adaptive operating region: mild compression can miss energy opportunities, and overly aggressive compression can hurt inference quality. Intermediate compression can reduce GPU energy by up to 53.2\%, and SoC energy by up to 48.2\%, with negligible quality loss. We argue for runtime policies that dynamically manage compression, guided by workload features and edge telemetry.

\end{abstract}


\maketitle

\section{Introduction}
\label{sec:intro}

Large language models (LLMs) have rapidly improved in capability and scale~\cite{kaplan2020scaling,brown2020language,ouyang2022training,touvron2023llama,rupprecht2026survey}, but they remain hampered by hallucinations~\cite{ji2023survey}, stale parametric knowledge~\cite{kasai2023realtime}, and poor access to private or domain-specific information~\cite{gao2023retrieval,lewis2020retrieval}. Retrieval-augmented generation (RAG) addresses these limitations by conditioning generation on passages retrieved at inference time, allowing the model to produce responses grounded in external, dynamic and domain-specific knowledge without retraining~\cite{lewis2020retrieval}. This makes RAG an attractive foundation for practical LLM applications in domains where correctness, freshness, and provenance matter.


Running RAG at the edge is compelling for privacy-sensitive data~\cite{li2024eacorag,ren2024ragmec}, latency-critical applications such as robotics~\cite{zhang2026retrieval,lan2025experience,lin2026vote,taherin2026glsvlsi} and augmented reality~\cite{dai2025eros,sun2020arvr}, personal agents~\cite{rawassizadeh2023odsearch,zheng2025review}, and mobile and bandwidth-limited settings where cloud access may be unreliable. Recent edge-RAG systems have improved efficiency on constrained platforms, including mobile devices~\cite{seemakhupt2024edgerag,park2025mobilerag}, edge SoCs~\cite{qin2024rocr}, and wearable systems~\cite{shao2025wearablerag}, through optimizations such as faster vector search, compact indexing, and reduced memory footprint. However, these systems primarily focus on retrieval/index efficiency or fixed content-reduction strategies.  Instead, we should explore how RAG can run efficiently on edge devices based on the application content.

This motivates the following questions: \emph{once documents have been retrieved, how much of that context should actually be passed to the generator on an edge device?} RAG improves knowledge access by adding retrieved tokens, but tokens are expensive on edge hardware. Each retrieved passage increases the prompt length seen by the generator, adding to prefill latency, KV-cache footprint, memory traffic, and energy per query~\cite{zheng2025review,seemakhupt2024edgerag}. This cost is amplified on constrained edge SoCs, where memory bandwidth, power budget, and thermal headroom are limited. Moreover, retrieving more context does not guarantee better answers: irrelevant or redundant passages can distract the generator and increase system cost without improving response quality~\cite{jiang2024longrag,liu2024lost,yoran2024making,taherin2018tsusc}. Thus, edge RAG comes with a new set of challenging tradeoff: \emph{the system needs enough context to preserve answer quality, but not so much context that generation overuses latency and energy.}

\begin{figure}[t]
\centering
\includegraphics[width=\columnwidth]{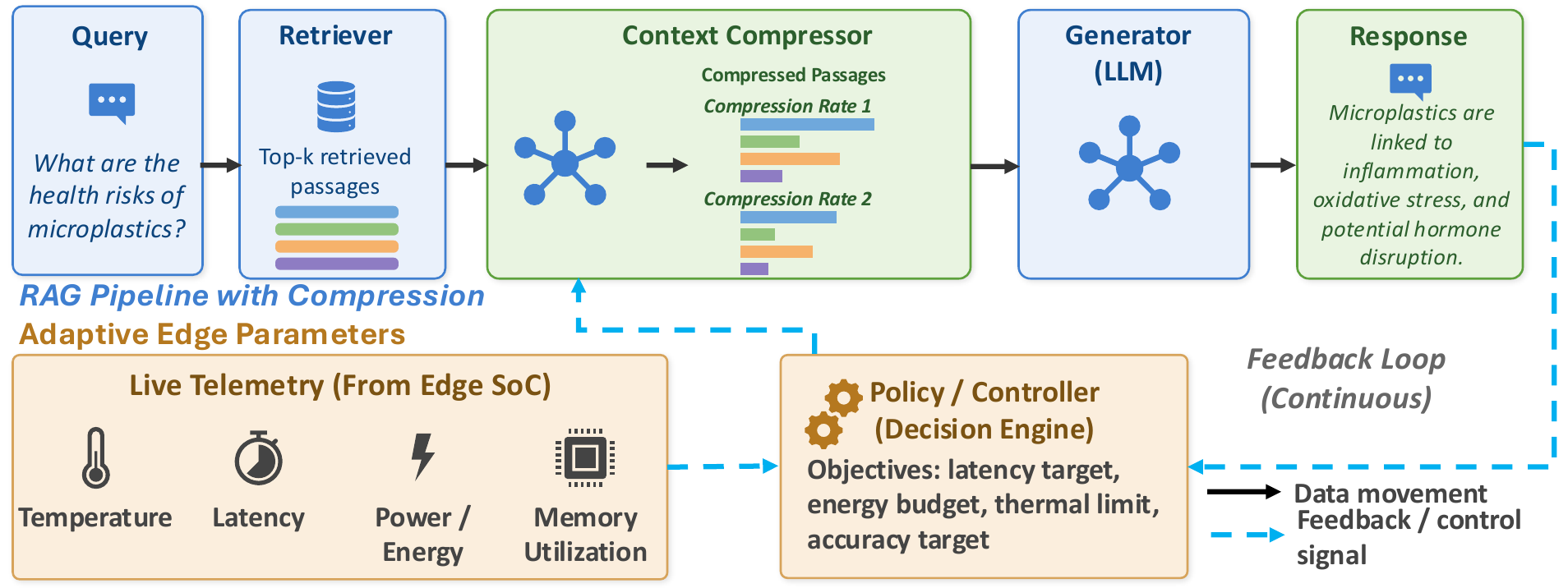}
\vspace{-0.3cm}
\hrule
\vspace{-0.35cm}
\caption{Telemetry-informed adaptive compression for edge RAG. Retrieved passages are compressed before generation, but the compression ratio is selected at runtime by a controller that observes edge-SoC telemetry and optimizes for latency, energy, thermal, memory, and accuracy constraints.}
\label{fig:adaptive-pipeline}
\vspace{-0.7cm}
\end{figure}

\emph{Context compression} is a natural knob for this tradeoff: it prunes or rewrites retrieved passages before generation, aiming to preserve query-relevant evidence while reducing input tokens. We use this term to distinguish our setting from broader prompt-compression work~\cite{li2025prompt}: only the retrieved context is compressed, while instructions and few-shot scaffolding remain unchanged. Unlike coarser controls such as retrieval depth, compression exposes a continuous rate knob: lower rates reduce generation load more aggressively, while higher rates preserve more evidence.


However, compression on the edge is not free. The compressor runs on the same SoC as retrieval and generation, adding latency, memory traffic, and energy while competing for shared resources. Thus, the relevant metric is not gross generation speedup but \emph{net benefit}: the generation savings from a shorter prompt minus the compressor's own overhead. A static compression ratio can be a net loss when contexts are short, generators are small, or too few tokens are removed.


In this paper, we present our vision for \emph{telemetry-informed adaptive compression}: context compression should be a runtime decision, not a fixed preprocessing step. As Figure~\ref{fig:adaptive-pipeline} illustrates, an edge RAG system could use workload features and live SoC telemetry (e.g., latency, energy, and memory bandwidth) to decide whether to compress and what rate to use. Our goal is to demonstrate the potential for such a controller by measuring when compression is beneficial, when it is harmful, and which system signals expose that boundary.

To evaluate this potential, we measure RAG on Jetson AGX Thor using Llama and Qwen generators from 1B to 8B parameters, Natural Questions and HotpotQA datasets, and LLMLingua-2 context compression. Our results show three findings. First, for 7B--8B models, generation dominates the edge-RAG budget, accounting for roughly 90\% of latency, 91\% of GPU energy, and 92\% of total SoC energy per query. Second, compression can provide substantial net savings after accounting for compressor overhead, but its benefit depends on model size, retrieval depth, and dataset. Third, sweeping the compression rate reveals an adaptive operating region: mild compression can lose energy, overly aggressive compression can hurt answer quality, and intermediate rates can substantially reduce energy (i.e., GPU energy by up to 53.2\% and SoC energy by up to 48.2\%) with little quality impact.

This paper makes three contributions: (1) a stage-level characterization of latency and energy in edge RAG on Jetson AGX Thor, identifying when generation becomes the dominant cost; (2) a net-benefit analysis of context compression that includes compressor overhead rather than reporting generation-only speedups; and (3) a telemetry-informed adaptive-compression vision, grounded in measured workload-dependent tradeoffs, for future edge-RAG systems.

\section{Background and Motivation}
\label{sec:background}

\subsection{RAG Pipeline and Context Cost}
\label{subsec:rag-pipeline}

A RAG pipeline has two phases: an offline indexing phase and an online query phase. Offline, a document collection is split into passages and stored in a searchable datastore, typically with lexical or vector indexes for efficient retrieval~\cite{robertson2009probabilistic,johnson2019billion}. Online, a user query is encoded and used to retrieve the top-ranked passages; a generator LLM then conditions on both the query and the retrieved context to produce the final response~\cite{lewis2020retrieval,izacard2020leveraging}. Modern RAG systems extend this basic retrieve-then-generate flow with additional stages such as query rewriting~\cite{ma2023query,wang2023query2doc}, reranking~\cite{nogueira2019passage,nogueira2020document}, and context compression~\cite{jiang2023llmlingua,li2025prompt}, inserted between retrieval and generation~\cite{gao2023retrieval,Gupta2024RAGsurvey}.

The same retrieved context that improves grounding also creates the main systems cost: every additional passage increases the prompt length seen by the generator. Longer prompts increase prefill work, KV-cache footprint and memory traffic, as well as the energy per query, which is especially costly on edge SoCs with limited memory bandwidth, power, and thermal headroom~\cite{zheng2025review,seemakhupt2024edgerag}. \emph{Context compression} is inserted between retrieval and generation to reduce this cost before the generator runs. Throughout this paper, we use the term \emph{rate} for the fraction of retrieved-context tokens \emph{kept}, matching the parameter exposed by LLMLingua-2~\cite{pan2024llmlingua}. Thus, $\text{rate}=0.5$ corresponds to keeping roughly half of the retrieved context, or about $2\times$ compression in factor notation. Compression is applied only to the retrieved context; the system prompt, instructions, and few-shot scaffolding (system message, few-shot
examples) are left intact.

\subsection{Context Compression Methods}
\label{subsec:compression-methods}
Prompt compression methods are commonly divided into \emph{hard} and \emph{soft} compression~\cite{li2025prompt}. Hard compression directly edits the textual input before inference, either by extracting important tokens or sentences or by summarizing the input~\cite{jiang2023llmlingua,pan2024llmlingua}. Soft compression instead maps text into latent tokens or embeddings consumed by the decoder~\cite{mu2023learning,ge2023context,chevalier2023adapting}. We focus on hard compression because it can be inserted into existing RAG pipelines without changing the generator interface and because compressed text length directly affects generation cost.

Hard compression can be \emph{extractive} or \emph{abstractive}. Extractive methods remove low-importance tokens or sentences while preserving the original text, as in the LLMLingua family~\cite{jiang2023llmlingua,pan2024llmlingua,jiang2024longllmlingua}. Abstractive methods such as RECOMP generate a new summary of the retrieved passages~\cite{xu2023recomp}, but this usually adds another autoregressive stage and therefore extra latency and energy. Prior device-oriented studies show that compressor cost varies substantially across methods~\cite{xu2024device,lu2024small}, making overhead a first-order concern on edge devices. Provence similarly shows the value of filtering context before generation, but does not expose the continuous rate knob needed for our adaptive-control study~\cite{chirkova2025provence}.

We use \textbf{LLMLingua-2}~\cite{pan2024llmlingua} as a representative compressor because it is extractive, lightweight relative to the generator, and exposes a continuous \emph{rate} parameter. This lets us sweep rate to measure the tradeoff among generation savings, compressor overhead, and answer quality. Our goal is not to claim LLMLingua-2 is the final edge-RAG compressor, but to use it as a controllable representative for identifying when compression helps, when it is a net loss, and where adaptive policies have room to operate.
\subsection{Edge RAG and the Missing Runtime View}
\label{subsec:edge-rag-gap}

Efficient edge inference has largely focused on reducing generator cost through quantization, memory reduction, KV-cache optimization, and runtime scheduling. RAG broadens this problem: retrieval, optional reranking or compression, and generation all share the same latency, memory, power, and thermal budget. Thus, a stage such as context compression must be evaluated by its \emph{net}  effect, not by generation speedup alone.

Recent edge-RAG systems address important parts of this pipeline. EdgeRAG optimizes retrieval memory and latency through index and embedding techniques~\cite{seemakhupt2024edgerag}; MobileRAG uses selective content reduction on mobile devices~\cite{park2025mobilerag}; RoCR accelerates retrieval with edge computing-in-memory architectures~\cite{qin2024rocr}; and wearable RAG systems optimize retrieval energy and memory movement~\cite{shao2025wearablerag}. This prior work shows that RAG can run on constrained platforms, but they primarily target retrieval/index efficiency, platform-specific acceleration, or fixed content reduction rather than runtime control of compression under live device constraints.

Adaptive RAG serving systems such as METIS show that RAG configurations should be selected per query rather than fixed offline~\cite{ray2025metis}. However, METIS targets server-side quality-delay tradeoffs, not edge SoC telemetry, energy, thermal behavior, or the cost of running a compressor on the same device as the generator. The missing runtime view is therefore deciding \emph{when compression is worth its own cost}: an edge controller must choose whether to compress, and at what rate, based on workload features, retrieved-context length, model size, quality risk, and hardware state.
\section{Methodology and Evaluation}
\label{sec:method}

\subsection{Experimental Setup and Metrics}
\label{subsec:setup}

We evaluate a \textbf{RAG pipeline} on the \textbf{NVIDIA Jetson AGX Thor}~\cite{nvidia_thor_trm_2025} using a standard retrieve-then-generate flow. Table~\ref{tab:rag-config} summarizes the shared configuration. All stages run on the same SoC, and the compressor, when enabled, is co-resident with the retriever and generator. We run two experiment groups: (1) an uncompressed stage-attribution sweep to identify where latency and energy are spent across model sizes, datasets, and retrieval depths, and (2) a compression-rate sweep to determine when LLMLingua-2 becomes net-positive and where energy savings begin to impact answer quality. The experimental setup builds on RAGMark, our stage-level RAG benchmarking framework~\cite{feric2026ragmark}, and incorporates the GPU/SoC power, thermal, and memory telemetry collection from our edge-LLM characterization framework, Hydra~\cite{taherin2026hydra}.

\begin{table}[t]
\centering
\caption{Experimental Setup}
\vspace{-0.3cm}
\label{tab:rag-config}
\scriptsize
\setlength{\tabcolsep}{3pt}
\renewcommand{\arraystretch}{0.88}
\begin{tabular}{p{0.17\columnwidth}p{0.82\columnwidth}}
\toprule
\textbf{Item} & \textbf{Configuration} \\
\midrule
Platform & Jetson AGX Thor: Blackwell GPU, 128\,GB LPDDR5x, 273\,GB/s, 130\,W. \\
Corpus / index & English Wikipedia 2018, sentence-split via FlashRAG ($\sim$9.4\,M passages)~\cite{jin2024flashrag}; e5-base-v2 encoder~\cite{intfloatE5baseV2}; FAISS GPU \texttt{IndexFlatL2} (~28.2\,GB). \\
QA Datasets & Natural Questions~\cite{kwiatkowski2019natural} and HotpotQA~\cite{yang2018hotpotqa}; 100 seed-paired queries per config. \\
Models & Llama-3.2 1B/3B, Llama-3.1 8B~\cite{grattafiori2024llama}; Qwen-2.5 1.5B/3B/7B~\cite{yang2024qwen2}. All fp16. \\
Compression & None, or LLMLingua-2~\cite{pan2024llmlingua} \\
Sweeps & Exp.~1 uses $k\in\{1,5,10\}$ with compression off. Exp.~2 uses HotpotQA, $k\in\{5,10\}$, and LLMLingua-2 rates 1.0, 0.9, 0.7, 0.5, 0.3, and 0.15. \\
Controls & Single-query mode, reranker off, standard pipeline, randomized config order, first 3 queries dropped as warm-up. \\
Metrics/telemetry & EM, token-level F1, retrieval recall; end-to-end and per-stage latency; GPU/SoC energy, power, memory, and temperature from \texttt{tegrastats} at 100\,ms cadence. \\
\bottomrule
\end{tabular}
\vspace{-0.4cm}
\end{table}

We report compression as a \emph{net} effect. Compressed RAG is beneficial only when the generation savings from a shorter prompt exceeds the compressor's own latency and energy on the same SoC. Therefore, our reported savings, include the compression stage itself, rather than treating generation-only speedup as an end-to-end gain.

\subsection{Results}
\label{subsec:results}


We answer three main questions: (i) where does the per-query budget go in an uncompressed pipeline, (ii) how do quality and net energy vary as we sweep the compression rate, and
(iii) how does the value of compression vary with workload characteristics?


\begin{figure}[t]
  \centering
  \includegraphics[width=\columnwidth]{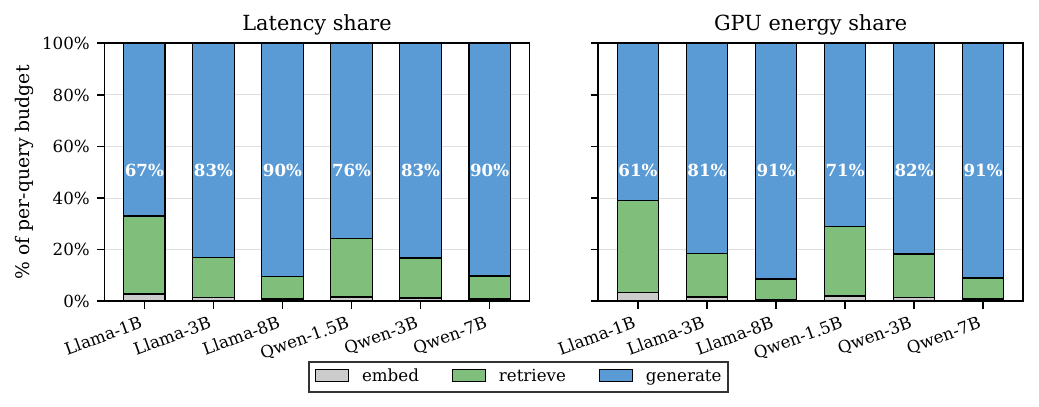}
  \vspace{-0.4cm}
  \hrule
  \vspace{-0.3cm}
  \caption{Per-query share of latency (left) and GPU energy (right) by stage, on AGX Thor, fp16, no compression.}
  \label{fig:stage-share}
  \vspace{-0.3cm}
\end{figure}

\textbf{Generation dominates the budget above 3\,B parameters
(Figure~\ref{fig:stage-share}).}
At Llama-3.1-8B and Qwen-2.5-7B, the generator alone accounts for
90\,\% of per-query latency and \emph{91\,\%} of per-query GPU
energy. Embed and retrieve together never exceed 10\,\% at this scale.
At the 1\,B end the picture inverts: Llama-3.2-1B spends 33\,\% of
wall time and 39\,\% of GPU energy in embed+retrieve, leaving
compression of the generator's prompt with much less headroom to
recover. The Llama-vs-Qwen split is statistically a no-op at equal
parameter counts (Llama-3B and Qwen-3B agree to within 0.3\,pp on the
generation share; Llama-8B and Qwen-7B to within 0.3\,pp), so we run
the compression sweeps on the Llama family only.

\begin{figure}[t]
  \centering
  \includegraphics[width=\columnwidth]{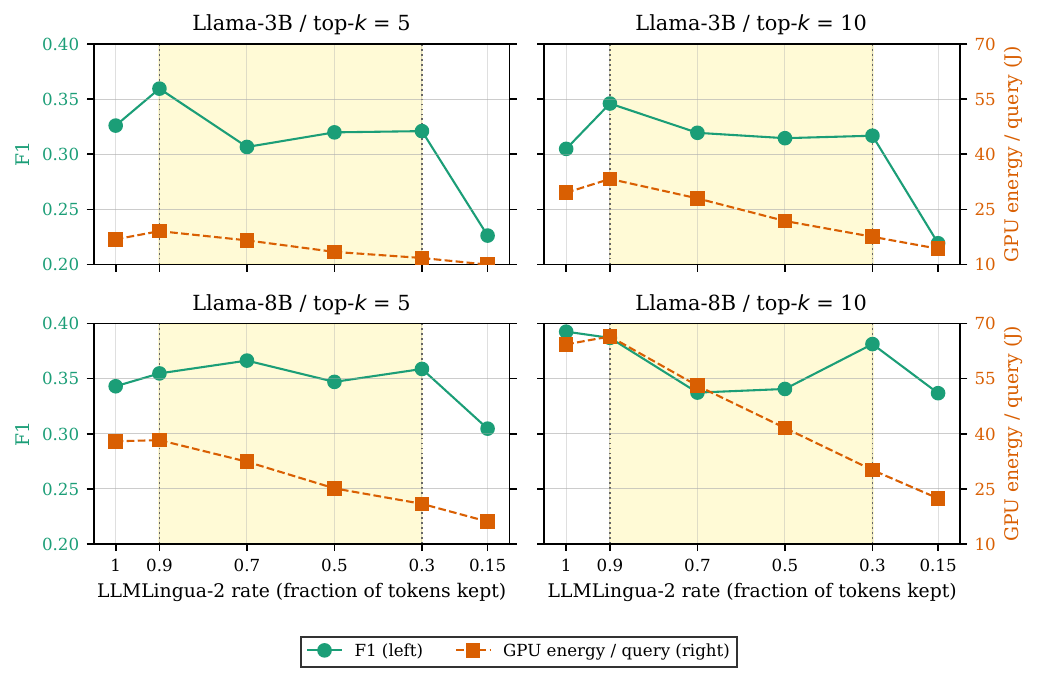}
  \vspace{-0.4cm}
  \hrule
  \vspace{-0.3cm}
  \caption{Answer F1 (left axis) and per-query GPU energy (right axis)
  vs.\ LLMLingua-2 $\text{rate}$ on HotpotQA. Shaded band: adaptive
  operating room between the two dotted-line knees.}
  \label{fig:knee}
  \vspace{-0.3cm}
\end{figure}

\textbf{Two knees frame a wide adaptive operating range
(Figure~\ref{fig:knee}).}
On every (model, top-$k$) panel, F1 is quite similar from
$\text{rate} = 1.0$ down to $\text{rate} = 0.3$ (the F1 deltas across
this range are inside the $\pm 0.05$ noise floor), then collapses by
4--10 absolute points at $\text{rate} = 0.15$ when the generator can no
longer reconstruct meaning from the heavily-pruned context. GPU energy changes monotonically: it gets \emph{worse} at $\text{rate} = 0.9$
because LLMLingua-2's roughly fixed $130$--$310$\,ms per-query
overhead is not amortized by dropping only 10\,\% of tokens; it gets
better from $\text{rate} = 0.7$ onward as the shorter prompt cuts
prefill cost. The two inflection points define an \emph{adaptive
operating region} of width $\approx 0.6$ on the rate axis (shaded). The
safest aggressive setpoint we observe is $\text{rate} = 0.3$,
which preserves quality and lies right at the quality knee.

\begin{figure}[t]
  \centering
  \includegraphics[width=0.85\columnwidth]{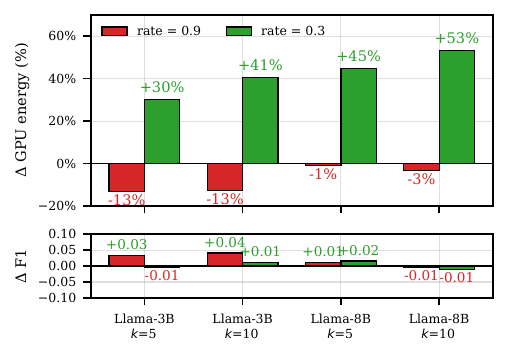}
  \vspace{-0.2cm}
  \hrule
  \vspace{-0.3cm}
  \caption{Net GPU energy delta (top) and $\Delta$F1 (bottom) at
  $\text{rate} = 0.9$ (mild) and $\text{rate} = 0.3$ (safe-aggressive),
  vs.\ the in-session uncompressed baseline.}
  \label{fig:savings}
  \vspace{-0.3cm}
\end{figure}

\textbf{The \emph{value} of compressing varies with workload, while the best observed safe-aggressive rate is stable (Figure~\ref{fig:savings},
Table~\ref{tab:headline-savings}).}
At $\text{rate} = 0.3$, GPU energy savings grow monotonically with
model size and retrieved context length, from $+30\,\%$ on Llama-3B
at $k{=}5$ to $+53\,\%$ on Llama-8B at $k{=}10$. At $\text{rate} = 0.9$
the relationship inverts: every configuration increases net energy usage (from
$-1\,\%$ on the heaviest workload to $-13\,\%$ on the lightest), because
the compressor's computational overhead outweighs the small generation
savings from dropping 10\,\% of tokens. The same $\text{rate}=0.3$
setpoint is the safest aggressive choice on every panel, while the
quantitative win it delivers shifts by a factor of $\sim$1.8$\times$
across our model+context combinations.

\begin{table}[t]
\centering
\caption{Net savings at $\text{rate} = 0.3$ vs.\ the in-session
uncompressed baseline (positive $\Delta$ = saving on the cost columns,
improvement on the quality columns).}
\vspace{-0.3cm}
\label{tab:headline-savings}
\setlength{\tabcolsep}{4pt}
\scriptsize
\begin{tabular}{lrrrrrr}
\toprule
Model & $k$ & $\Delta$lat. & $\Delta E_{\text{GPU}}$ & $\Delta E_{\text{SoC}}$ & $\Delta$EM & $\Delta$F1 \\
\midrule
Llama-3B & 5  &  $+$8.5\% & $+$30.2\% & $+$28.2\% & $+$0.000 & $-$0.005 \\
Llama-3B & 10 & $+$15.6\% & $+$40.6\% & $+$35.1\% & $+$0.010 & $+$0.012 \\
Llama-8B & 5  & $+$25.2\% & $+$44.9\% & $+$38.5\% & $+$0.052 & $+$0.016 \\
\textbf{Llama-8B} & \textbf{10} & \textbf{+32.6\%} & \textbf{+53.2\%} & \textbf{+48.2\%} & \textbf{+0.000} & \textbf{$-$0.011} \\
\bottomrule
\end{tabular}
\vspace{-0.2cm}
\end{table}

\section{Discussion and Research Agenda}
\label{sec:discussion}

\subsection{From Adaptive Room to an Online Controller}

The two-knee structure in Figure~\ref{fig:knee} reframes adaptive compression on edge devices. In our sweep, the best observed safe-aggressive rate is $\text{rate}=0.3$, but the \emph{value} of enabling compression varies substantially with workload. Heavy workloads, such as 8\,B generators with $k{=}10$, recover up to 53\% of GPU energy at this setpoint, while lighter workloads recover only 30\%. Mild compression at $\text{rate}=0.9$ is a net loss because the compressor's fixed per-query cost is not amortized by dropping only a small fraction of tokens.

This suggests that the first practical controller does not need to search a continuous rate space for every query. Instead, it can make a small number of workload-conditioned decisions: skip compression when projected generation savings fall below compressor overhead, use $\text{rate}=0.3$ for long-context or high-cost generations, and reserve more aggressive compression for settings with explicit quality slack. Such a controller can use cheap signals already available at runtime, including model identity, retrieved-context length, top-$k$, recent latency, energy, and thermal headroom. The solution is therefore not only choosing a compression method, but deciding when compression is worth its own cost on the same SoC.

\subsection{Research Agenda}

Three questions naturally arise. First, \textbf{how does the energy knee move with lighter compressors?} LLMLingua-2 uses a fixed model pass, making mild compression expensive. Streaming, token-level, or hardware-aware compressors could make compression beneficial at milder rates and shrink the negative-savings region.

Second, \textbf{do the observed knees hold under quantization and larger models?} Our measurements use fp16 generators up to 8\,B parameters. Production edge deployments often use 4-bit quantization, and larger models may shift the balance between generation savings, compressor overhead, and quality loss.

Third, \textbf{how should compression interact with the rest of the RAG pipeline?} Compression is only one knob. Retrieval depth, reranker cutoff, source selection, query rewriting, and conditional or iterative RAG policies~\cite{asai2024self} all consume the same latency, memory, energy, and thermal budget. An telemetry-informed controller, that considers multiple RAG knobs, should lead to significant savings. 

Treating compression as an adaptive on-device resource-management knob, rather than a fixed preprocessing step, makes efficient grounded generation on the edge an interesting optimization problem. Our two-knee structure, evaluated on AGX Thor, offers empirical evidence for building an adaptive compression controller.

\section{Conclusion}
This paper presents a vision, grounded in empirical evidence, that context compression should be treated as a runtime systems knob for edge RAG rather than a fixed preprocessing step. Our measurements on Jetson AGX Thor show why: generation dominates the cost of edge-RAG pipelines, compression can deliver substantial net energy savings, and the benefit depends on workload and compression rate. The observed gap between the energy and quality knees demonstrates the potential for telemetry-informed adaptive compression. Future edge-RAG systems should use workload features and live SoC state to decide when compression is worth its own cost and how aggressively it should be applied.

\bibliographystyle{ACM-Reference-Format}
\bibliography{references}

\end{document}